\documentclass[conference]{IEEEtran}
\IEEEoverridecommandlockouts
\usepackage{cite}
\usepackage{amsmath,amssymb,amsfonts}
\usepackage{algorithmic}
\usepackage{graphicx}
\usepackage{textcomp}
\usepackage{xcolor}

\usepackage{multirow}
\usepackage{adjustbox}
\usepackage{siunitx}
\usepackage{graphicx}
\usepackage{booktabs}
\usepackage{array}
\usepackage{float}
\usepackage{spconf,amsmath,graphicx}
\usepackage{stfloats}
\usepackage{color}
\usepackage{hyperref}

\def\BibTeX{{\rm B\kern-.05em{\sc i\kern-.025em b}\kern-.08em
    T\kern-.1667em\lower.7ex\hbox{E}\kern-.125emX}}
\begin{document}

\title{EHNet: An Efficient Hybrid Network for Crowd Counting and Localization}

\name{Yuqing Yan, Yirui Wu*}

\address{Hohai University}

\maketitle

\begin{abstract}
In recent years, crowd counting and localization have become crucial techniques in computer vision, with applications spanning various domains.
The presence of multi-scale crowd distributions within a single image remains a fundamental challenge in crowd counting tasks.
To address these challenges, we introduce the Efficient Hybrid Network (EHNet), a novel framework for efficient crowd counting and localization.
By reformulating crowd counting into a point regression framework, EHNet leverages the Spatial-Position Attention Module (SPAM) to capture comprehensive spatial contexts and long-range dependencies. Additionally, we develop an Adaptive Feature Aggregation Module (AFAM) to effectively fuse and harmonize multi-scale feature representations. Building upon these, we introduce the Multi-Scale Attentive Decoder (MSAD).
Experimental results on four benchmark datasets demonstrate that EHNet achieves competitive performance with reduced computational overhead, outperforming existing methods on ShanghaiTech Part \_A, ShanghaiTech Part \_B, UCF-CC-50, and UCF-QNRF. Our code is in \href{https://anonymous.4open.science/r/EHNet}{https://anonymous.4open.science/r/EHNet}.
\end{abstract}


\section{Introduction}
\label{sec:intro}

Crowd counting, aimed at accurately quantifying large gatherings of individuals, has emerged as a critical research area at the intersection of artificial intelligence and computer vision.
In recent years, the applications of crowd counting technology have expanded significantly, encompassing diverse domains such as public safety and urban planning \cite{CLTR}, traffic management \cite{FGENet}, video surveillance \cite{P2PNet}, and smart city initiatives \cite{HGNN}.
While previous research has established fundamental counting capabilities, significant challenges persist, particularly in scenarios involving high crowd density and high occlusion.
While previous research has established fundamental counting capabilities, significant challenges persist, particularly in scenarios involving high crowd density and high occlusion. The presence of substantial scale variations among individuals in crowded scenes, caused by high density and occlusion, demands methods with both strong long-range feature capturing ability and fine-grained local information extraction capabilities.
To address these challenges, researchers have proposed various deep learning architectures, each with distinct approaches to feature extraction and scale handling.

Neural network architectures for crowd counting can be broadly categorized into two main approaches: Convolutional Neural Network (CNN) based approaches \cite{DSNet,SANet} and Transformer-based approaches \cite{CLTR,Vit}. CNN-based approaches typically take images or video frames as input and generate either crowd density maps or direct count estimates. Notable examples in this category include 
MCNN \cite{MCNN}, CSRNet \cite{CSRNet}, and  FGENet \cite{FGENet}.
Although CNNs excel at capturing fine-grained local patterns and hierarchical spatial features, their inherent architectural constraints in modeling long-range dependencies significantly hinder their capability to extract discriminative features for large-scale individuals.

\begin{figure}
    \centering
    \includegraphics[width=1\linewidth]{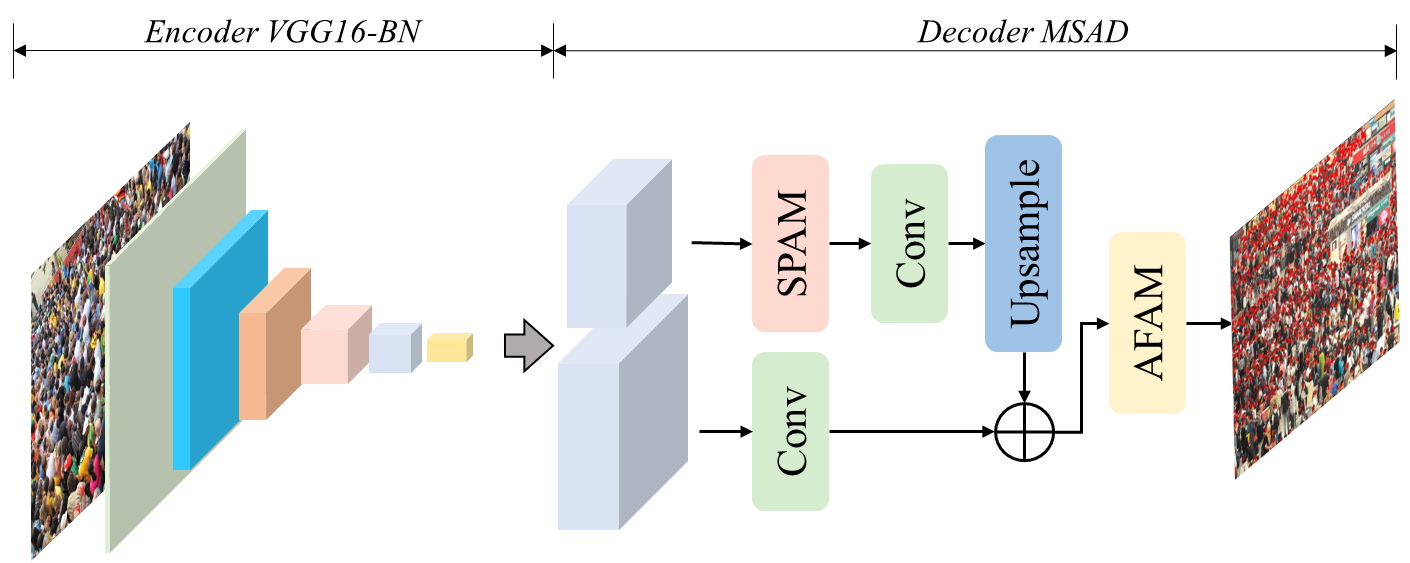}
    \caption{Overall architecture diagram of our model. }
    \label{fig:model}
\end{figure}

In contrast, Transformer-based approaches, inspired by Vision Transformer (ViT) \cite{Vit}, offer an alternative to CNNs in crowd counting by processing image patches to produce density maps or count estimates. 
While Transformer-based models such as CLTR \cite{CLTR} and LoViTCrowd \cite{LoViT} demonstrate superior capability in modeling long-range spatial dependencies, they are fundamentally constrained by two critical limitations: inefficient extraction of fine-grained local features and substantial computational complexity.
The limitations of existing approaches motivate us to develop a novel architecture that efficiently integrates global dependency modeling with local feature extraction. While Transformer-based methods excel at capturing global dependencies, their quadratic computational complexity severely limits their practicality in high-resolution crowd counting scenarios. To address this challenge, we propose an Efficient Hybrid Network (EHNet) that achieves global dependency modeling with linear computational complexity while preserving the fine-grained feature extraction capabilities of CNNs. The detailed structure of EHNet is illustrated in Fig.\ref{fig:model}.
This framework combines global contextual modeling with fine-grained feature extraction, specifically designed to address the challenges of multi-scale variations in crowd scenes. Specifically, our approach leverages a Spatial-Position Attention Module (SPAM) attention mechanism to extract comprehensive global dependency, while incorporating fine-grained details through a Adaptive Feature Aggregation Module (AFAM).

Our contributions can be summarized as follows:
\begin{itemize}
\item We propose an end-to-end hybrid network named EHNet. EHNet demonstrates superior performance in both accurate crowd counting and precise localization of individuals.

\item We propose three innovative modules: the SPAM attention mechanism, which effectively captures global dependency while preserving spatial features; the AFAM module, which enhances the network's capacity to extract and process fine-grained information for improved counting accuracy; and the MSAD, which builds upon the SPAM and AFAM components to further enhance the model's performance.

\item We demonstrate the effectiveness of our approach through extensive experiments.
\end{itemize}
\section{Our approaches}
\label{sec:approaches}
\subsection{Point framework}
In point framework, the model directly utilizes the original annotation points in the image as prediction targets, where the annotation points precisely indicate the actual position of each individual. Suppose there are $\mathnormal{N}$ individuals in an image, the position of the $\mathnormal{i}$-th individual is represented by the coordinate pair ($\mathnormal{x}_i$, $\mathnormal{y}_i$), where \(i \in \{1, 2, \dots, N\}\), with $\mathnormal{x}_i$ as the horizontal coordinate and $\mathnormal{y}_i$ as the vertical coordinate.

During the inference process, given an input image, the model first generates a series of anchor points through a feature extraction network, with a stride of 2 pixels. Then, a regression head is used to predict the offset of each anchor point relative to the true individual's position, as well as the corresponding confidence score. These offsets and confidence scores reflect the model's prediction of the position of the individual associated with each anchor point. To ensure prediction reliability, we filter out low-confidence predictions using a threshold $\tau$. Suppose the predicted number of individuals after thresholding is $\mathnormal{M}$, the predicted point coordinates are represented by $(\hat{x}_j, \hat{y}_j)$, where $j \in \{1, 2, \dots, M\}$, denoting the position of the $\mathnormal{j}$-th predicted point.

During training, the model needs to match the annotation points with the predicted points to calculate the position loss. Ideally, each predicted point ($\hat{x}_j$, $\hat{y}_j$) should be as close as possible to the corresponding annotation point ($\mathnormal{x}_i$, $\mathnormal{y}_i$), meaning the model should accurately predict and align with the annotated point’s location. This matching process can be formalized as a bipartite matching problem, where one set of nodes consists of the annotation points and the other set consists of the predicted points. The Hungarian algorithm is used to optimize the matching between the predicted and annotated points, ensuring that the cost of matching, typically represented as the distance between the annotation and predicted points, is minimized.

\subsection{Efficient hybrid network}
This model design adopts an encoder-decoder architecture, primarily aimed at crowd counting and localization tasks. In this architecture, the model first uses a pre-trained VGG16-bn \cite{vgg16} as the encoder to extract multi-scale features from the input image. 

The extracted multi-scale features are then fed into a module called the Multi-Scale Attentive Decoder. The purpose of this decoder is to progressively decode the feature maps, further refining and enhancing the fine-grained information of individuals in the crowd. The output of the decoding process mainly consists of two components: offset values and confidence scores. The offset values are used to predict the displacement of each anchor point relative to the center of the actual individual, while the confidence scores indicate the likelihood that the anchor point has detected an individual. By combining these two outputs, the model can not only effectively estimate the total number of individuals in the image but also accurately locate each individual in dense crowd scenes.



\begin{figure}
    \centering
    \includegraphics[width=0.9\linewidth]{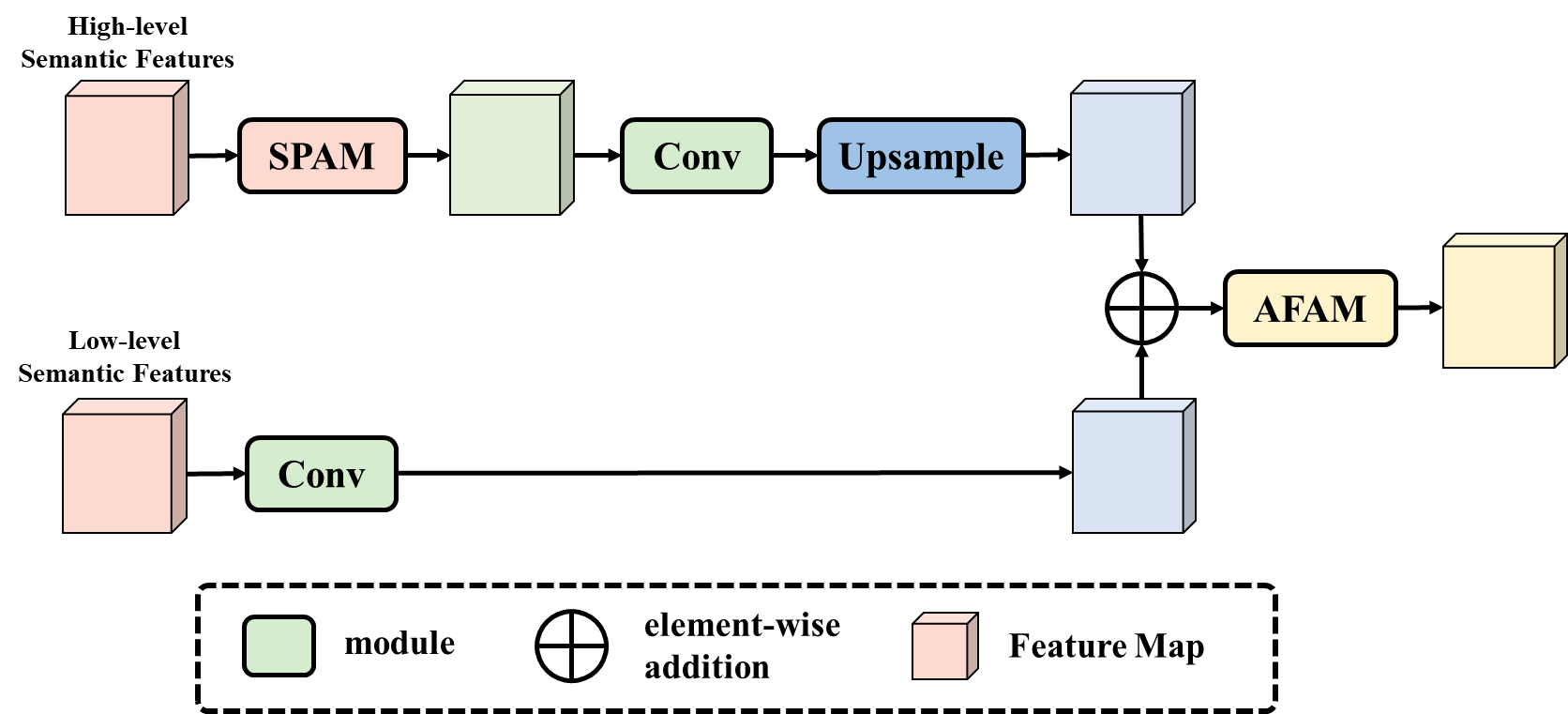}
    \caption{The architecture of MSAD. }
    \label{fig:decoder}
\end{figure}

\subsection{Multi-Scale Attentive Decoder}

Our model decoder consists of two main modules: Spatial-Position Attention Module (SPAM) and Adaptive Feature Aggregation Module (AFAM), shown in Fig.\ref{fig:decoder}.

\begin{figure}[!h]
    \centering
    \includegraphics[width=0.8\linewidth]{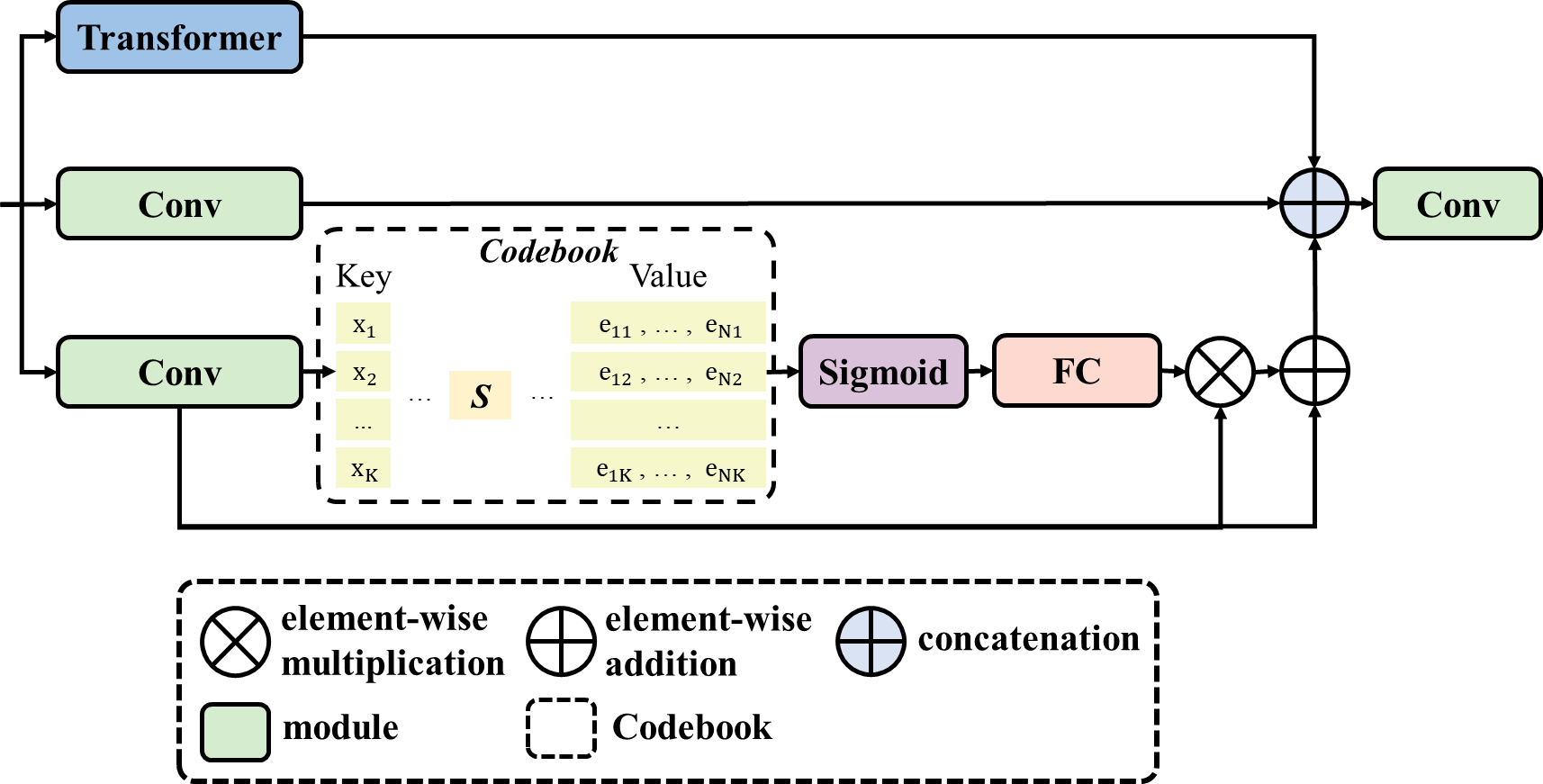}
    \caption{The architecture of SPAM. }
    \label{fig:SPAM}
\end{figure}

\textbf{Spatial-Position Attention Module (SPAM).} SPAM is a novel three-branch architecture that synergizes Transformer Block, convolution operations, and Cipherbook Module, as illustrated in Fig.\ref{fig:SPAM}. This design enables comprehensive multi-scale feature learning, effectively capturing both spatial dependencies and position-sensitive patterns in dense crowd scenes.

Each branch in SPAM serves a distinct yet complementary purpose: the Transformer Block models long-range dependencies for global context understanding, the convolution branch processes local spatial features with position sensitivity, and the Cipherbook Module captures explicit spatial patterns through a learned dictionary of feature representations.  Through this synergistic combination, SPAM enhances the model's ability to accurately count and locate individuals in diverse crowd scenarios.

The three-branch architecture operates simultaneously on different aspects of the input features, enabling SPAM to process global context, local details, and spatial structures in parallel. This parallel processing strategy not only improves computational efficiency but also provides a more comprehensive feature representation, where each branch's unique strengths contribute to the overall performance in crowd counting and localization tasks.
\begin{figure}[!h]
    \centering
    \includegraphics[width=\columnwidth]{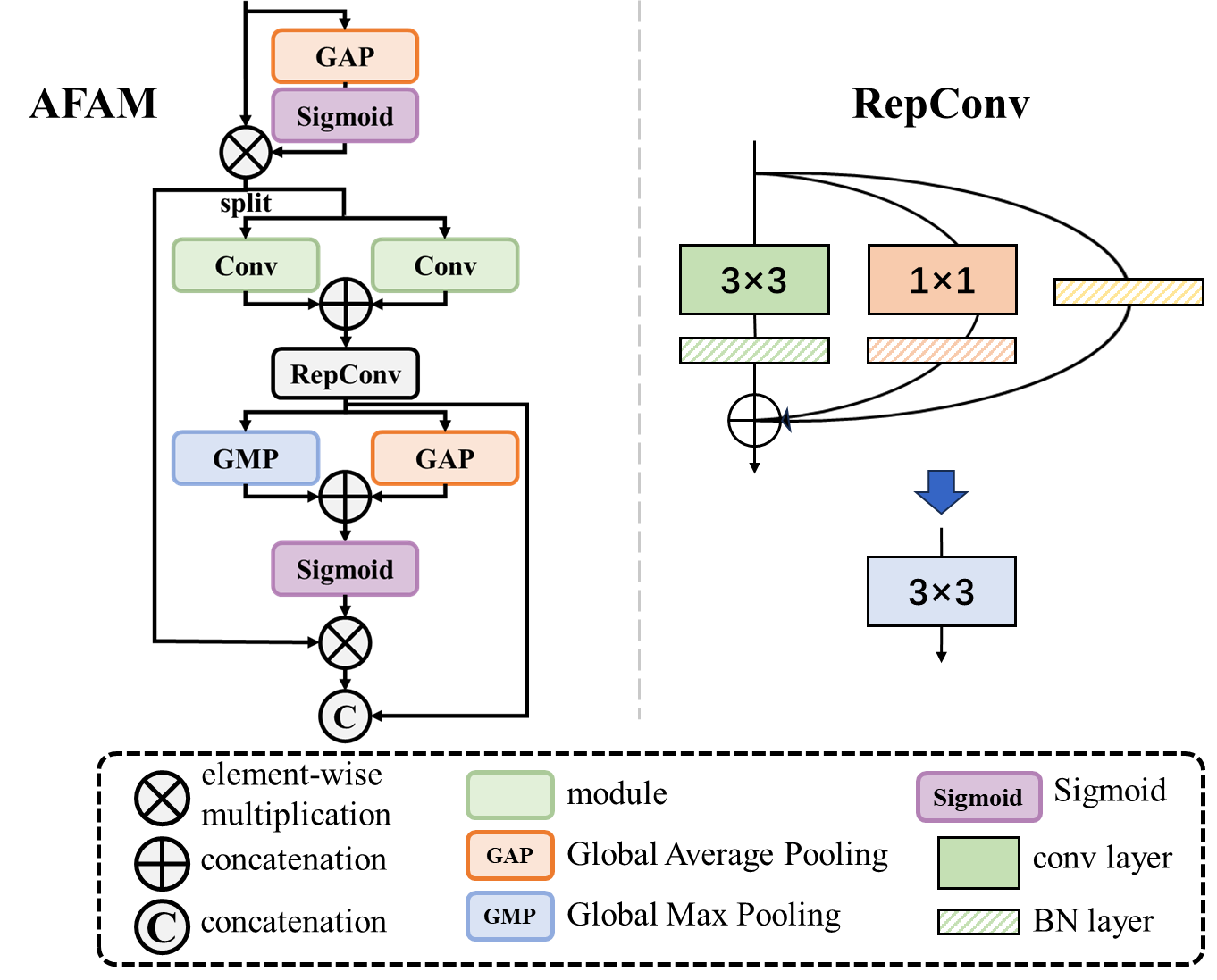}
    \caption{The architecture of AFAM. }
    \label{fig:AFAM}
\end{figure}

\textbf{Adaptive Feature Aggregation Module (AFAM).} AFAM is an efficient feature aggregation network that combines channel-wise attention with multi-scale feature learning, as shown in Fig.\ref{fig:AFAM}. The module integrates parallel convolutions with RepConv \cite{repvgg} for feature extraction, while employing both channel and spatial attention mechanisms to dynamically emphasize informative features. This dual-attention design, coupled with the inference-optimized RepConv structure, enables AFAM to achieve both computational efficiency and effective feature representation.

The module first compresses spatial information to obtain channel descriptors, which are transformed into weights through a Sigmoid function to enhance important information across different channels.

The weighted features are then split into two paths. The first path processes features through two parallel convolution layers, whose outputs are concatenated and fed into a RepConv module. As illustrated in Fig.\ref{fig:AFAM}, RepConv consists of parallel 3×3 and 1×1 convolutions with BN layers, which can be fused into a single 3×3 convolution during inference. The second path applies both Global Max Pooling (GMP) and Global Average Pooling (GAP) operations, combines their outputs through addition, and generates spatial attention weights via a Sigmoid function. These weights are applied to the features through element-wise multiplication.

Finally, the module concatenates the outputs from both paths to produce the final aggregated features, effectively combining multi-scale spatial information with attention-enhanced representations.

\begin{table*}[!t]
\centering
\caption{Comparison of different approaches on four datasets.}
\label{tab:comparison}
\resizebox{0.75\textwidth}{!}{%
\begin{tabular}{l c cccc cccc}
\toprule
\multirow{2}{*}{\textbf{Approaches}} & \multirow{2}{*}{\textbf{Venue}} & \multicolumn{4}{c}{\textbf{ShanghaiTech}} & \multicolumn{2}{c}{\textbf{UCF\_CC\_50}} & \multicolumn{2}{c}{\textbf{UCF-QNRF}} \\
\cmidrule(lr){3-6} \cmidrule(lr){7-8} \cmidrule(lr){9-10}
& & \multicolumn{2}{c}{Part A} & \multicolumn{2}{c}{Part B} & & & & \\
& & MAE & MSE & MAE & MSE & MAE & MSE & MAE & MSE \\
\midrule
MCNN \cite{MCNN} & CVPR'16 & 110.2 & -- & 26.4 & -- & 377.6 & -- & -- & -- \\
CAN \cite{CANNet} & CVPR'19 & 62.8 & 101.8 & 7.7 & 12.7 & 212.2 & 243.7 & 107.0 & 183.0 \\
ASNet \cite{ASNet} & CVPR'20 & 57.78 & 90.13 & -- & -- & 174.84 & 251.63 & 91.59 & 159.71 \\
SUA \cite{SUA} & ICCV'21 & 68.5 & 121.9 & 14.1 & 20.6 & -- & -- & 130.3 & 226.3 \\
GauNet \cite{GauNet} & CVPR'22 & 54.8 & 89.1 & \underline{6.2} & 9.9 & 186.3 & 256.5 & 81.6 & 153.7 \\
HDNet \cite{HDNet} & ICME'22 & 53.4 & 89.9 & -- & -- & -- & -- & 83.2 & 148.3 \\
CLTR \cite{CLTR} & ECCV'23 & 56.9 & 95.2 & 6.5 & 10.6 & -- & -- & 85.8 & 141.3 \\
DDC \cite{DDC} & CVPR'23 & 52.87 & 85.62 & 6.08 & \underline{9.61} & 157.12 & 220.59 & \textbf{65.79} & \textbf{126.53} \\
CHS-Net \cite{CHS-Net} & ICASSP'23 & 59.2 & 97.8 & 7.1 & 12.1 & -- & -- & 83.4 & 144.9 \\
FGENet \cite{FGENet} & MMM'23 & \underline{51.66} & \underline{85.0} & 6.34 & 10.53 & \underline{142.56} & \underline{215.87} & 82.1 & 143.76 \\
STEERER \cite{STEERER}  & ICCV'23 & 54.5 & 86.9 & \textbf{5.8} & \textbf{8.5} & -- & -- & 79.5 & 144.3 \\
VMambaCC \cite{VMambaCC} & ArXiv'24 & 51.87 & \textbf{81.3} & 7.48 & 12.47 & -- & -- & 88.42 & 144.73 \\
Gramformer \cite{Gramformer} & AAAI'24 & 54.7 & 87.1 & -- & -- & -- & -- & \underline{76.7} & \underline{129.5} \\
M2PLNet \cite{M2PLNet} & ICME'24 & \textbf{50.86} & 89.86 & -- & -- & \textbf{123.3} & \textbf{185.14} & -- & -- \\
\midrule
\textbf{Ours} & & \textbf{50.29} & \textbf{81.01} & \textbf{6.64} & \textbf{10.46} & \textbf{136.2} & \textbf{211.37} & \textbf{78.25} & \textbf{132.63} \\
\bottomrule
\end{tabular}%
}
\end{table*}

\begin{table*}[!t]
\centering
\caption{Efficiency of different approaches.}
\label{tab:efficiency}
\resizebox{0.75\textwidth}{!}{%
\begin{tabular}{l c cccc cccc}
\toprule
\multirow{2}{*}{\textbf{Approaches}} & \multicolumn{2}{c}{\textbf{Parameters}} & \multicolumn{2}{c}{\textbf{Computation}} & \multicolumn{2}{c}{\textbf{Memory Usage}} & \multicolumn{2}{c}{\textbf{Self Time}} \\
\cmidrule(lr){2-3} \cmidrule(lr){4-5} \cmidrule(lr){6-7} \cmidrule(lr){8-9}
& Total & Trainable & FLOPs & Average Inference Time & GPU & CPU & CPU & CUDA \\
\midrule
MCNN \cite{MCNN} & 133,705 & 133,705 & 2.692G & 1.88 ms & 33.59 MB & 1994.00 MB & 11.692ms & 11.777ms \\
CSRNet \cite{CSRNet} & 16,263,489 & 16,263,489 & 41.445G & 5.53 ms & 74.12 MB & 1996.07 MB & 6.089ms & 8.128ms \\
CAN \cite{CANNet} & 16,530,113 & 16,530,113 & 41.868G & 6.61 ms & 82.57 MB & 2333.90 MB & 12.297ms & 12.306ms \\
ASNet \cite{ASNet} & 30,398,087 & 30,398,087 & 62.745G & 9.34 ms & 0.57 MB & 2294.65 MB & 28.600ms & 28.623ms \\
CLTR \cite{CLTR} & - & - & - & - & OOM & - & - & - \\
FGENet \cite{FGENet} & 183,271,139 & 183,271,139 & 64.647G & 22.14 ms & 280.14 MB & 2413.40 MB & 74.955ms & 74.964ms \\
STEERER \cite{STEERER} & 39,296 & 39,296 & 3.969G & 1.34 ms & 61.83 MB & 2334.20 MB & 4.787ms & 5.466ms \\
Gramformer \cite{Gramformer} & -- & -- & 60.9G & 12.6 ms & 0.57 MB & 2286.12 MB & -- & -- \\
M2PLNet \cite{M2PLNet} & 508,460,865 & 508,460,865 & 234.103G & 59.99 ms & 369.08 MB & 2420.55 MB & 178.544ms & 178.531ms \\
\midrule
\textbf{Ours} & \textbf{2,515,008} & \textbf{2,515,008} & \textbf{133.871G} & \textbf{11.08 ms} & \textbf{161.57 MB} & \textbf{2287.59 MB} & \textbf{5.314ms} & \textbf{16.510ms} \\
\bottomrule
\end{tabular}%
}
\end{table*}

\section{Experiments}
\subsection{ Datasets and implementation details}
\textbf{Datasets:}
To comprehensively evaluate our model's performance, we utilized four representative crowd counting datasets: ShanghaiTech Part A, ShanghaiTech Part B, UCF CC 50, and UCF-QNRF. These datasets cover a wide range of crowd densities and scenarios, providing a robust testing framework. ShanghaiTech Part A contains 482 high-density images (average 501 individuals per image), while Part B includes 716 lower-density street scenes (average 123 individuals). UCF CC 50, despite having only 50 images, presents extreme variations in crowd counts (94-4,543 individuals). UCF-QNRF, a large-scale dataset, comprises 1,535 high-resolution images with over 1.3 million annotated individuals across diverse settings.

\textbf{Implementation details:}
The experiments were conducted on a computing platform equipped with an RTX 4060 TI GPU with 16GB of memory. To enhance the model's generalization ability and robustness, we implemented data augmentation techniques, including random flipping and random cropping. For optimization, we employed the Adam optimizer with an initial learning rate of 1e-4.

\subsection{Experiment}
\textbf{Counting experiment:}
Experimental results in Tab.\ref{tab:comparison} demonstrate our method's strong performance across four benchmark datasets. On ShanghaiTech Part A, we achieve state-of-the-art results with MAE/MSE of 50.29/81.01, outperforming recent approaches like M2PLNet and VMambaCC. While STEERER leads on ShanghaiTech Part B, our method shows competitive performance with MAE/MSE of 6.64/10.46. For UCF\_CC\_50, we achieve the second-best results (136.2/211.37) after FGENet. On UCF-QNRF, our approach maintains competitive performance (78.25/132.63) compared to leading methods DDC and Gramformer. These results demonstrate our method's strong generalization ability across diverse crowd counting scenarios.

\textbf{Efficiency experiment:}
Our approach demonstrates significant advantages across multiple dimensions, shown in Tab.\ref{tab:efficiency}.
The parameters are substantially lower than that of other approaches, indicating a more lightweight model. This is highly beneficial for deployment in resource-constrained environments, while also enhancing efficiency during both training and inference processes.
Although our FLOPs is relatively higher than other approaches, the model achieves outstanding performance through reduced inference time and highly efficient resource utilization, suggesting superior computational efficiency.
Memory usage on both GPU and CPU remains within a reasonable range, making the model well-suited for practical engineering deployment.
While the self CUDA time is higher compared to some lightweight approaches, such as STEERER, the self CPU time demonstrates significantly greater computational acceleration efficiency compared to approaches like ASNet and CAN.
Overall, our approach exhibits notable advantages in terms of parameters, memory usage, and self CPU time. It is a lightweight, high-performance, and computationally efficient crowd counting approach, making it particularly suitable for real-time applications and deployment scenarios.
\begin{table}
\centering
\caption{Localization experimental results on SHT\_A}
\begin{adjustbox}{max width=0.75\columnwidth}
    
\begin{tabular}{lccc}
\hline
Approaches & P(\%) & R(\%) & F(\%) \\
\hline
LCFCN \cite{LCFCN} & 43.30 & 26.00 & 32.50 \\
Method in \cite{Methodin}& 34.90 & 20.70 & 25.90 \\
LSC-CNN \cite{LSC-CNN}& 33.40 & 31.90 & 32.60 \\
TopoCount \cite{TopoCount}& 41.70 & 40.60 & 41.10 \\
CLTR \cite{CLTR}& 43.60 & 42.70 & 43.20 \\
P2PNet \cite{P2PNet}& 46.00 & 41.50 & 39.80 \\
\textbf{Ours} & \textbf{52.78} & \textbf{51.53} & \textbf{52.13} \\
\hline
\end{tabular}
\end{adjustbox}
\label{tab:localization_results}
\end{table}

\textbf{Localization experiment:}
Experimental results demonstrate that our approach achieves state-of-the-art performance in crowd localization on the SHT\_A dataset, as shown in Tab.\ref{tab:localization_results}. Using Hungarian algorithm for point matching with a distance threshold of 4 pixels to determine positive and negative samples, our method achieves 52.78\% precision, 51.53\% recall, and 52.13\% F1 score. This represents a significant improvement of approximately 9 percentage points across all metrics compared to the previous best method P2PNet. The balanced improvement in both precision and recall metrics demonstrates our method's robust performance in accurate crowd localization.
\begin{figure*}
    \centering
    \includegraphics[width=0.85\linewidth]{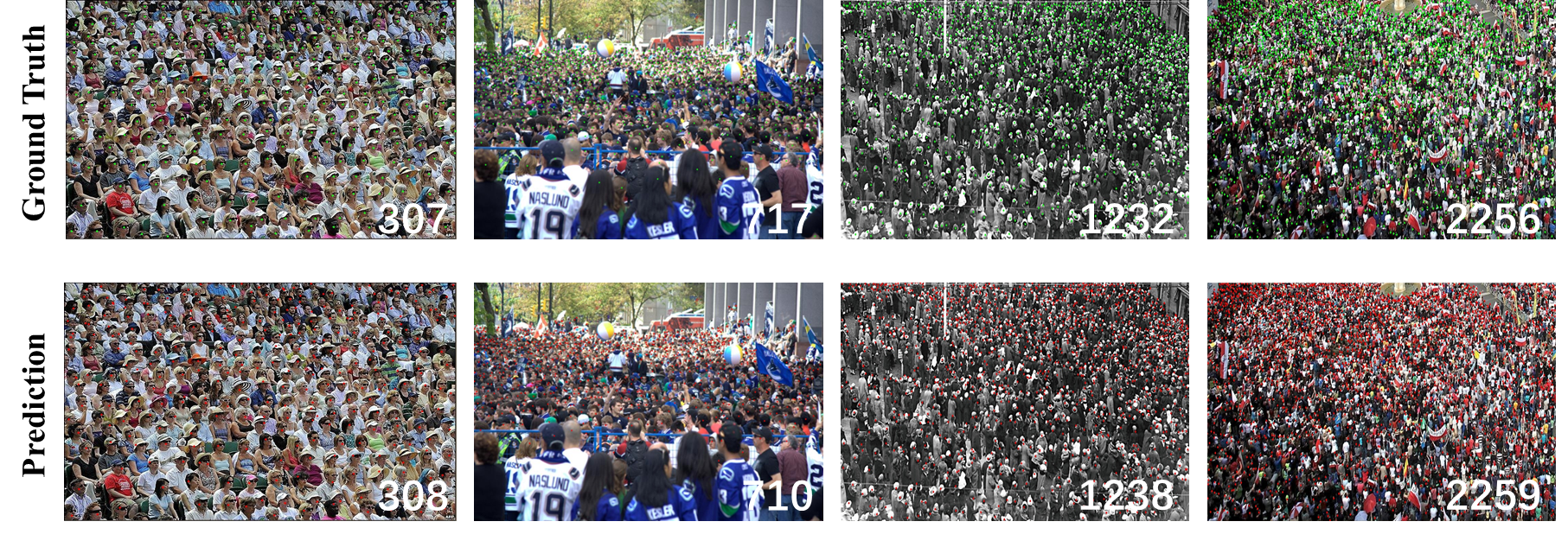}
    \caption{The predicted results of our EHNet. The white numbers denote to the ground truth or prediction number.}
    \label{fig:visualization}
\end{figure*}
\subsection{Ablation experiment :}
\begin{table}[!t]
\centering
\caption{Model ablation experiment on SHT\_A}
\label{tab:model}
\begin{adjustbox}{max width=\columnwidth}
\begin{tabular}{lccc}
\toprule
\textbf{Module Name} & \textbf{MAE} & \textbf{MSE} \\
\midrule
BaseLine & 54.02 & 82.59 \\
+SPAM & 51.28 & 81.93 \\
+AFAM & 50.97 & 82.64 \\
\textbf{Ours} & \textbf{50.29} & \textbf{81.01} \\
\bottomrule
\end{tabular}
\end{adjustbox}
\end{table}

The ablation study of model demonstrates that the complete model configuration (Ours) achieved the best performance across both MAE and MSE, shown in Tab.\ref{tab:model}. Adding the SPAM module reduced the MAE to 51.28, a decrease of 2.74, and slightly improved the MSE to 81.93. Incorporating the AFAM module further reduced the MAE to 50.97, showing a 3.05 improvement compared to the baseline, while the MSE increased slightly to 82.64. Combining all modules together achieved the lowest MAE of 50.29 and the lowest MSE of 81.01. Compared to the baseline, the MAE decreased by approximately 3.73, and the MSE improved by 1.58. The superior performance of ours highlights that our full model configuration achieves the most accurate and robust results, outperforming the baseline and other partial configurations.

\begin{table}[!h]
\centering
\caption{AFAM ablation experiment on SHT\_A}
\label{tab:module_performance}
\begin{adjustbox}{max width=\columnwidth}
\begin{tabular}{lccc}
\toprule
\textbf{Module Name} & \textbf{MAE} & \textbf{MSE} \\
\midrule
Conv & 55.46 & 92.27 \\
ELAN \cite{ELAN} & 54.91 & 86.36 \\
CSPLayer \cite{CSPLayer} & 53.78 & 87.48 \\
\textbf{AFAM} & \textbf{50.29} & \textbf{81.01} \\
\bottomrule
\end{tabular}
\end{adjustbox}
\end{table}

The AFAM module achieved the best performance on both metrics, with an MAE of 50.29 and an MSE of 81.01, shown in Tab.\ref{tab:module_performance}. Compared to ELAN and CSPLayer, AFAM reduced the MAE by approximately 4.62 and 3.49 respectively, and decreased the MSE by 5.35 and 6.47 respectively. These results demonstrate that the AFAM module has significant advantages in improving counting accuracy and reducing errors, confirming its effectiveness in crowd counting tasks.

\begin{table}[!h]
\centering
\caption{SPAM ablation experiment on SHT\_A}
\label{tab:module_comparison}
\begin{adjustbox}{max width=0.85\columnwidth}
\begin{tabular}{cllcc}
\toprule
\textbf{Exp. No.} & \textbf{Module} & \textbf{Configuration} & \textbf{MAE} & \textbf{MSE} \\
\midrule
1 & Conv & Baseline & 54.02 & 82.59 \\
2 & {+Transformer} & Conv + Transformer & 54.00 & 84.32 \\
3 & {+CodeBook} & Conv + CodeBook & 52.46 & 81.58 \\
4 & {\begin{tabular}[c]{@{}l@{}}+Transformer \\ \& CodeBook\end{tabular}} & \begin{tabular}[c]{@{}l@{}}Conv + Transformer \\ + CodeBook\end{tabular} & \textbf{50.29} & \textbf{81.01} \\
\bottomrule
\end{tabular}
\end{adjustbox}
\end{table}

This ablation study compared the performance of different module combinations, shown in Tab.\ref{tab:module_comparison}. The results show that the complete model configuration (Conv + Transformer + CodeBook) achieved the best performance in both MAE and MSE, with values of 50.29 and 81.01 respectively. Compared to the baseline model, MAE decreased by approximately 3.73, while MSE reduced by 1.58. Adding either the Transformer or CodeBook module alone led to performance improvements, but the combination of both yielded the best results. This indicates that the Transformer and CodeBook modules have a synergistic effect in improving crowd counting accuracy, confirming the effectiveness of model design.

\section{Conclusion}
In this work, we propose a novel end-to-end hybrid architecture. To address the multi-scale issue caused by image perspective effects, we introduce SPAM, AFAM and MSAD. Through extensive experiments, we demonstrate the effectiveness of our approach in crowd localization and counting, achieving impressive results across multiple datasets.

\bibliographystyle{IEEEbib}

\end{document}